\documentclass[letterpaper, 10 pt, conference]{ieeeconf}  

\IEEEoverridecommandlockouts                              

\usepackage{graphicx} 
\usepackage{epsfig} 
\usepackage{amsmath} 
\usepackage{amssymb}  
\usepackage{tabularx}
\usepackage{multirow}
\usepackage{makecell}
\usepackage{subcaption}
\usepackage{cite}
\usepackage{siunitx}

\DeclareMathOperator*{\argmax}{argmax} 
\usepackage[absolute,showboxes]{textpos}

\TPMargin*{3pt}
\newcommand\copyrighttext{
    \footnotesize
    \noindent
    SUBMITTED TO REVIEW AND POSSIBLE PUBLICATION. COPYRIGHT WILL BE TRANSFERRED WITHOUT NOTICE.\\
    Personal use of this material is permitted.
    Permission must be obtained for all other uses, in any current or future media, including reprinting/republishing this material for advertising or promotional purposes, creating new collective works, for resale or redistribution to servers or lists, or reuse of any copyrighted component of this work in other works.}%
\newcommand\copyrightnotice{%
    \begin{textblock*}{6.6in}(0.95in,0.15in)
        \centering
        \copyrighttext
    \end{textblock*}
}

\title{\LARGE \bf
Pioneering SE(2)-Equivariant Trajectory Planning\\for Automated Driving
}

\author{Steffen Hagedorn$^{1}$, Marcel Milich$^{2}$, and Alexandru P. Condurache$^{1}$
\thanks{$^{1}$Robert Bosch GmbH, 71229 Leonberg, Germany and Institute for Signal Processing, Universität zu Lübeck, 23562 Lübeck, Germany. {\tt \scriptsize steffen.hagedorn@de.bosch.com},~
$^{2}$Bosch Center for Artificial Intelligence, 71272 Renningen, Germany and Institute for Parallel Distributed Systems, Universität Stuttgart, 70569 Stuttgart, Germany}
}

\begin{document}
\copyrightnotice

\maketitle
\thispagestyle{empty}
\pagestyle{empty}

\begin{abstract}
Planning the trajectory of the controlled ego vehicle is a key challenge in automated driving.
As for human drivers, predicting the motions of surrounding vehicles is important to plan the own actions.
Recent motion prediction methods utilize equivariant neural networks to exploit geometric symmetries in the scene.
However, no existing method combines motion prediction and trajectory planning in a joint step while guaranteeing equivariance under roto-translations of the input space.
We address this gap by proposing a lightweight equivariant planning model that generates multi-modal joint predictions for all vehicles and selects one mode as the ego plan.
The equivariant network design improves sample efficiency, guarantees output stability, and reduces model parameters.
We further propose equivariant route attraction to guide the ego vehicle along a high-level route provided by an off-the-shelf GPS navigation system.
This module creates a momentum from embedded vehicle positions toward the route in latent space while keeping the equivariance property.
Route attraction enables goal-oriented behavior without forcing the vehicle to stick to the exact route.
We conduct experiments on the challenging nuScenes dataset to investigate the capability of our planner.
The results show that the planned trajectory is stable under roto-translations of the input scene which demonstrates the equivariance of our model.
Despite using only a small split of the dataset for training, our method improves L2 distance at \qty{3}{s} by \qty{20.6}{\%} and surpasses the state of the art.

\end{abstract}

\section{INTRODUCTION}

\begin{figure}[tpb]
    \centering
    \includegraphics[width=\columnwidth]{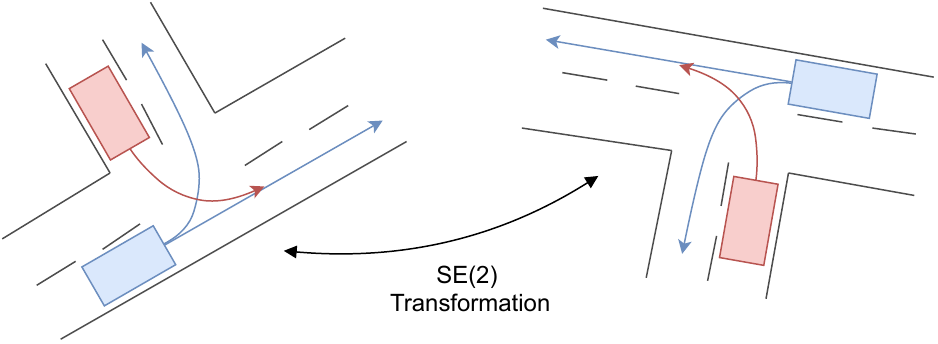}
    \caption{Exemplary traffic scene that demonstrates the intuition behind SE(2)-equivariant trajectory prediction and planning: Roto-translations of the input scene should result in an equivalent transformation of the trajectory output.}
    \label{traffic_scene_bev}
\end{figure}

In automated driving, trajectory planning is the task of finding a safe and efficient path and speed profile of a controlled ego vehicle (EV) toward a goal position \cite{hallgarten2023prediction}.
In addition to past positions, map, and route information, many planning methods rely on motion prediction of surrounding vehicles (SVs) to model interactions~\cite{rhinehart2019precog,song2020pip,hallgarten2023prediction,hu2023planning}.
Combining prediction and planning by handling all vehicles jointly is a promising approach to reduce computation and overcome purely reactive behavior~\cite{rhinehart2019precog, huang2023gameformer, hagedorn2023rethinking}.
To increase sample efficiency and robustness, the predicted trajectories of all vehicles must be independent of the viewpoint from which the scene is observed (cf. Fig.~\ref{traffic_scene_bev})\cite{cui2023gorela}.
Equivariant models fulfill this requirement and are therefore utilized in many tasks that solve problems in observable physical systems~\cite{satorras2021n, xu2023eqmotion, walters2020trajectory}.
Equivariance means that transformations of the input transform the output in an equivalent way.
Designing equivariant neural networks (NNs) is beneficial in multiple ways.
Beside guaranteeing output stability they have an increased sample efficiency and can reduce model parameters~\cite{rath2022improving}.
When given a fixed-size dataset, inducing prior knowledge by means of equivariance can increase the performance~\cite{rath2020boosting}.

While methods for joint prediction and planning have been presented and the advantages of equivariance have been used in stand-alone motion prediction, no existing planning model combines both techniques.\\
Instead, many methods transform the whole scene into a coordinate system centered around one vehicle, typically the ego vehicle~\cite{ngiam2021scene,gilles2021thomas,zhai2023rethinking}.
However, this approach has proven sample-inefficient and vulnerable to domain shifts as the scene representation is viewpoint-dependent~\cite{cui2023gorela, hallgarten2023stay}.
Other works alleviate this problem by taking the perspective of each vehicle~\cite{gao2020vectornet,cui2019multimodal,janjovs2022starnet}.
Such methods are more robust but computationally expensive~\cite{kim2022stopnet} which also is a known downside of universally applicable equivariant approaches that are based on irreducible representations of the transformation group~\cite{weiler20183d,weiler2019general}.
In contrast, EqMotion reduces the computation by explicitly designed equivariant operations that do not rely on irreducible representations~\cite{xu2023eqmotion}.
EqDrive applies EqMotion for vehicle motion prediction in traffic scenes~\cite{wang2023eqdrive}.
Their results demonstrate that equivariant neural networks can improve the performance in automated driving tasks.

Since the benefits of joining prediction and planning as well as the advantages of equivariant models are shown in recent methods, we want to pioneer the field of combining both aspects.
Therefore we propose PEP (Pioneering Equivariant Planning): A lightweight equivariant planning method that integrates prediction and planning in a joint approach.
Similar to EqMotion~\cite{xu2023eqmotion}, PEP is a graph neural network~\cite{kipf2016semi} that consists of an equivariant feature learning branch that processes vehicle positions and an invariant branch that processes invariant features such as the distance between two positions.
We extend the architecture by adding another equivariant branch that updates the EV position by providing route information.
This allows the joint processing of EV and SVs while conditioning the EV prediction on a goal.
We further add a mode selection layer that selects the EV plan from $K$ predicted modes.
The network is trained with a loss that jointly optimizes planning and prediction, and promotes diverse multi-modal predictions.

We evaluate PEP on the prediction split of nuScenes~\cite{caesar2020nuscenes}.
Alongside the open-loop evaluation, we present a comprehensive ablation study and equivariance analysis.

In summary, our contributions are:
\begin{itemize}
    \item We present the first equivariant planner integrated with multi-modal joint prediction.
    \item We propose an equivariant route attraction mechanism that allows following a high-level route.
    \item We report state-of-the-art performance on the nuScenes dataset in open-loop planning.
\end{itemize}

\section{RELATED WORK}

\subsection{Joint Prediction and Planning}

Early planning models for automated driving plan the EV's trajectory directly from perception inputs without explicitly considering the interplay with SVs~\cite{hagedorn2023rethinking}.
Alternatively, many solutions sequentially employ separate subsystems for prediction and planning~\cite{cui2021lookout, song2020pip, dauner2023parting, chen2023tree}.
While increasing explainability, such methods still handle EV and SVs separately and lead to reactive behavior~\cite{hagedorn2023rethinking}.
By modeling the future of all vehicles simultaneously, joint prediction and planning goes beyond reactive behavior and can reduce computation~\cite{rhinehart2019precog,huang2023gameformer}. Joint prediction and planning approaches can be categorized into iterative methods and regression.\\
The iterative probabilistic method PRECOG predicts the state of all vehicles one step into the future and uses the outcome as input for the next iteration~\cite{rhinehart2019precog}.
Goal information is provided for the EV and leads to more precise predictions for all vehicles.
The EV's plan is then inferred by optimizing the expectation of the predicted distribution.
GameFormer is another iterative approach based on game theory~\cite{huang2023gameformer}.
Interactions are modeled as a level-$k$ game in which the individual predictions are updated based on the predicted behavior of all vehicles from the previous level.
The encoder-decoder transformer architecture predicts multiple modes for SVs while restricting EV prediction to a single mode that serves as the plan.\\
In contrast, regressive methods learn a joint feature for the whole prediction horizon from which complete trajectories are regressed.
SafePathNet employs a transformer for multi-modal joint prediction of EV and SVs~\cite{pini2022safe}.
Every predicted EV mode is then checked for collisions with the most probable mode of each SV.
The EV mode with the lowest predicted collision rate is selected as the plan.
Similarly, DIPP starts with a multi-modal joint prediction and selects the mode with the highest probability for each vehicle~\cite{huang2022differentiable}.
To infer the EV plan, a differentiable nonlinear optimizer refines the EV prediction under consideration of the SV predictions and additional hand-crafted constraints.\\
We also base our planner on multi-modal joint prediction for all vehicles but further design the whole network to be equivariant under 2D roto-translations of the input.

\subsection{Equivariant Motion Prediction}

Equivariant convolutional neural networks add rotation equivariance to the inherent translation equivariance of the convolution operation~\cite{cohen2016group}.
Rotation equivariance in the 2D image domain is achieved by oriented convolutional filters~\cite{marcos2017rotation}, log-polar coordinates~\cite{esteves2017polar}, circular harmonics~\cite{worrall2017harmonic} or steerable filters~\cite{weiler2018learning,rath2020boosting}.
With the advent of Graph Neural Networks (GNNs)~\cite{kipf2016semi} which work on sparse data representations, equivariant adaptions of this architecture emerged.
To extract roto-translation-equivariant features from point clouds or graphs, some approaches utilize irreducible representations of the transformation group such as spherical harmonics~\cite{thomas2018tensor,fuchs2020se}.
Others base their method on specifically designed and computationally less expensive layers~\cite{jing2020learning,deng2021vector,kofinas2021roto}. For example, \cite{deng2021vector} embed the representation of neurons into a 3D space where they leverage basic principles of matrix algebra to induce equivariance.\\
Equivariant GNNs are a common choice for solving tasks in physical systems since these often possess rotational and translational symmetries~\cite{ummenhofer2019lagrangian,sanchez2020learning}.
Since motion prediction models a physical system, equivariant NNs are well-suited for this task.
\cite{wang2020incorporating}~first apply learning-based equivariant motion prediction to predict fluid flow.
SE(3)-equivariant transformers mark another milestone by regressing pedestrian trajectories~\cite{fuchs2020se}. 
The first equivariant motion prediction model for autonomous driving utilizes polar coordinate-indexed filters to design an equivariant continuous convolution operator~\cite{walters2020trajectory}.
Recently, the motion prediction network EqMotion presents strong performance on various tasks~\cite{xu2023eqmotion}.
Similar to~\cite{deng2021vector}, its specifically designed layers exploit geometrical symmetries.
In addition to the equivariant feature learning they present an invariant interaction reasoning module and an invariant pattern feature.
This design allows to integrate prior knowledge efficiently.
Features like absolute distances which are inherently SE(3)-invariant can be processed in the invariant layers while absolute positions are handled by the SE(3)-equivariant layers.
EqDrive finally applies EqMotion for vehicle trajectory prediction~\cite{xu2023eqmotion}.
In this work, we propose an equivariant model based on EqMotion that extends the motion prediction to a trajectory planner for the EV and integrates prior knowledge of its intended route.

\section{METHOD}
\begin{figure*}[tpb]
    \centering
    \includegraphics[width=\linewidth]{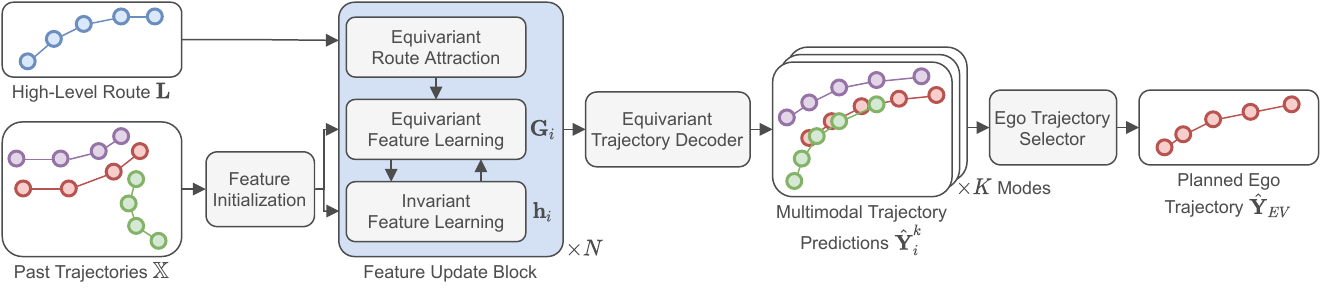}
    \caption{PEP model overview. After feature initialization, the features are updated N times in three parallel but interacting branches. A multi-modal decoder then predicts multiple future scenarios for all vehicles jointly. Alongside the trajectories, a probability for each scenario is estimated. The EV trajectory of the most probable mode is selected as the plan.}
    \label{model}
\end{figure*}

In this section, we introduce our equivariant trajectory planner PEP, which is an expansion of EqMotion.
For a more detailed overview of EqMotion, we refer the reader to~\cite{xu2023eqmotion}.
Fig.~\ref{model} provides an overview of our model.

\subsection{Problem Formulation}
PEP is an equivariant trajectory planner based on multi-modal joint prediction of all vehicles and trained by imitation learning.
Given the past trajectories $\textbf{X}_i = [x_i^1, x_i^2, \dots, x_i^{T_p}] \in \mathbb{R}^{T_p \times 2}$ of $i = 1, \dots, M$ vehicles, including the EV,
and EV route information $\textbf{L} \in \mathbb{R}^{C \times 2}$,
the planning task is to forecast $\hat{\textbf{Y}}_{EV} = [\hat{y}_{EV}^1, \hat{y}_{EV}^2, \dots, \hat{y}_{EV}^{T_f}] \in \mathbb{R}^{T_f \times 2}$
as close to the real future trajectory $\textbf{Y}_{EV}$ as possible.
We further denote the set of all past trajectories as $\mathbb{X} = [\textbf{X}_1, \dots, \textbf{X}_M]$.\\
Especially, we require the planning function $\mathcal{F}_{\text{plan}}(\mathbb{X},L) = \hat{\textbf{Y}}_{EV}$ to be equivariant under transformations $\mathcal{T}_g$ in the Euclidean group SE(2), which comprises rotations $\textbf{R} \in SO(2)$ and translations $\textbf{t} \in \mathbb{R}^2$.
All feature updates $f$ in $\mathcal{F}_{\text{plan}}$ must satisfy the equivariance condition $f(x\mathcal{T}_g) = f(x)\mathcal{T}_g$
where the roto-translation right group action $\mathcal{T}_g$ acts on 2D inputs $x$ via matrix-multiplication and addition $x\mathcal{T}_g = x\textbf{R} + \textbf{t}$.

\subsection{Feature Initialization}
The key idea of handling a set of positions translation-equivariantly is to shift the viewpoint into the center, i.e. the mean coordinate $\bar{\mathbb{X}}$.
To return to the initial coordinate system after a transformation, $\bar{\mathbb{X}}$ is re-added.
Like EqMotion~\cite{xu2023eqmotion}, we initialize the equivariant feature of vehicle $i$ as
\begin{equation}
\label{eq:equiv_feature_init}
    \textbf{G}_i^{(0)} = \phi_{\text{init}_g}(\textbf{X}_i - \bar{\mathbb{X}}) + \bar{\mathbb{X}} \in \mathbb{R}^{C\times 2}
\end{equation}
where function $\phi_{\text{init}_g}$ is realized by a fully connected layer (FCL)~\cite{xu2023eqmotion}.
In the following, all $\phi$ describe FCLs.
Since an FCL is a linear transformation and can be expressed as a matrix multiplication, rotation-equivariance follows from the multiplicative associative law.
We initialize the invariant feature of vehicle $i$ as a function of velocity $\Delta \textbf{X}_i$ and heading angle, which are both inferred from positions $\textbf{X}_i$ as in EqMotion~\cite{xu2023eqmotion}.
The $[\cdot ; \cdot]$ operator denotes concatenation.
\begin{equation}
    \textbf{h}_i^{(0)} = \phi_{\text{init}_h}([||\Delta \textbf{X}_i||_2 ; \text{angle}(\Delta \textbf{X}_i^{\tau}, \Delta \textbf{X}_i^{\tau-1})]) \in \mathbb{R}^D
\end{equation}
EqMotion further adds an invariant relationship learning, which computes $\textbf{c}_{ij} \in [0,1]^Q$ between agents $i$ and $j$ from the initial equivariant and invariant feature~\cite{xu2023eqmotion}.
$\textbf{c}_{ij}$ describes the relationship of $i$ and $j$ in $Q$ categories.
For instance, the network could learn to extract distance, velocity differences or heading differences of $i$ and $j$.

\subsection{Feature Update}

\subsubsection{Equivariant Route Attraction}
Many automated driving systems comprise a tactical planner or navigation system to provide a coarse intended route at lane level.
We introduce a novel module called 'equivariant route attraction' to incorporate the intended EV route into the joint prediction.
The intuition is to move the equivariant feature of the EV toward the high-level route $\textbf{L}$ in latent space before considering interactions with other vehicles.
This order of feature updates prioritizes social interactions over route following, which is important for collision avoidance.
Since the goal is only known for the EV ($i=0$), we update only this feature:
\begin{equation}
\label{eq:route_att_update_rule}
f_{\text{ra}} : \textbf{G}_0^{(l)} \leftarrow \textbf{G}_0^{(l)} + \phi_{\text{ra}}^{(l)} ( \textbf{L} - \textbf{G}_0^{(l)} ) \in \mathbb{R}^{C \times 2}.
\end{equation}
The FCL $\phi_{ra}$ takes vector $\textbf{L} - \textbf{G}_0^{(l)}$ as input, which points from the equivariant EV feature embedding toward the route.
Superscript $(l)$ denotes the $l$-th of $N$ feature update blocks.
We show that the route attraction module $f_{\text{ra}}$ fulfills the equivariance condition stated in the problem formulation:
\begin{equation}
\begin{split}
f_{\text{ra}}(x\textbf{R}+\textbf{t})
& = \textbf{G}_0^{(l)}\textbf{R} + \textbf{t} + \phi_{\text{ra}} ( \textbf{L}\textbf{R} + \textbf{t} - (\textbf{G}_0^{(l)}\textbf{R} + \textbf{t}) )\\
& = \textbf{G}_0^{(l)}\textbf{R} + \textbf{t} + \phi_{\text{ra}} ( (\textbf{L} - \textbf{G}_0^{(l)})\textbf{R} )\\
& = \textbf{G}_0^{(l)}\textbf{R} + \textbf{t} + \phi_{\text{ra}} ( \textbf{L} - \textbf{G}_0^{(l)} )\textbf{R}\\
& = (\textbf{G}_0^{(l)} + \phi_{\text{ra}} ( \textbf{L} - \textbf{G}_0^{(l)} ) )\textbf{R} + \textbf{t}\\
& = f_{\text{ra}}(x)\textbf{R}+\textbf{t}_{\quad \square}
\end{split}
\end{equation}

\subsubsection{Equivariant Feature Learning}
This feature update step comprises inner aggregation and neighbor aggregation.
Inner aggregation updates the equivariant feature of vehicle $i$ using a weight computed from its invariant feature. $\bar{\mathbb{G}}^{(l)}$ is the mean position of equivariant features $\textbf{G}_i^{(l)}$~\cite{xu2023eqmotion}:
\begin{equation}
    \textbf{G}_i^{(l)} \leftarrow \phi_{\text{att}}^{(l)}(\textbf{h}_i^{(l)}) \cdot (\textbf{G}_i^{(l)} - \bar{\mathbb{G}}^{(l)}) + \bar{\mathbb{G}}^{(l)} \in \mathbb{R}^{C \times 2}
\end{equation}
Neighbor aggregation first defines an edge weight for each neighbor based on relationship feature $\textbf{c}_{ij}$, equivariant, and invariant features.
The $i$-th equivariant feature is then updated by a weighted sum over all its neighbors $\mathcal{N}_i$~\cite{xu2023eqmotion}.
\begin{equation}
    \textbf{e}_{ij}^{(l)} = \sum_{k=1}^K \textbf{c}_{ij,k} \phi_{e,k}^{(l)}([\textbf{h}_i^{(l)};\textbf{h}_j^{(l)};||\textbf{G}_i^{(l)} - \textbf{G}_j^{(l)}||_2]) \in \mathbb{R}^C
\end{equation}
\begin{equation}
    \textbf{G}_i^{(l)} \leftarrow \textbf{G}_i^{(l)} + \sum_{j \in \mathcal{N}_i} \textbf{e}_{ij}^{(l)} \cdot (\textbf{G}_i^{(l)} - \textbf{G}_j^{(l)}) \in \mathbb{R}^{C \times 2}
\end{equation}
Finally, we apply the equivariant non-linear function proposed in~\cite{xu2023eqmotion} to infer $\textbf{G}_i^{(l+1)}$.

\subsubsection{Invariant Feature Learning}
The last step of the feature update in EqMotion~\cite{xu2023eqmotion} is invariant feature learning:
\begin{equation}
    \textbf{p}_i^{(l)} = \sum_{j \in \mathcal{N}_i} \phi_m^{(l)} ([\textbf{h}_i^{(l)} ; \textbf{h}_j^{(l)} ; ||\textbf{G}_i^{(l)} - \textbf{G}_j^{(l)}||_2]) \in \mathbb{R}^D
\end{equation}
\begin{equation}
    \textbf{h}_i^{(l+1)} = \phi_\textbf{h}^{(l)} ([\textbf{h}_i^{(l)} ; \textbf{p}_i^{(l)}]) \in \mathbb{R}^D
\end{equation}

\subsection{Trajectory Decoding}
To achieve multi-modal predictions, we introduce $K$ parallel FCL trajectory decoders.
Each decoder predicts all agents simulateneously based on their equivariant features:
\begin{equation}
    \hat{\textbf{Y}}_i^k = \phi_{\text{dec}}^k (\textbf{G}_i^N - \bar{\mathbb{G}}^N) + \bar{\mathbb{G}}^N \in \mathbb{R}^{(T_f+1) \times 2}
\end{equation}
Note that we predict an additional output beyond prediction horizon $T_f$.
It serves as a probability indicator for the trajectory selector, which outputs the final EV plan.

\subsection{Trajectory Selection}
We define mode probability as the mean of the spatial coordinate dimension $C$ of the additionally predicted point.
\begin{equation}
    \textbf{P}_i^k = \text{mean}_C(\hat{\textbf{Y}}_i^{k, T_f+1}) \in \mathbb{R}^K
\end{equation}
Selecting the most probable mode yields the EV plan:
\begin{equation}
    \hat{\textbf{Y}}_{EV} = \hat{\textbf{Y}}_0^{k^\ast} \quad \text{where} \quad k^\ast = \argmax_{k=1, \dots, K} \textbf{P}_0^k
\end{equation}
To promote mode diversity we apply a winner-takes-all (WTA) loss as described below.

\subsection{Training Objective}
In accordance with the problem statement, we focus on the planning performance in the loss function.
Additionally, prediction performance for SVs is optimized in order to benefit from realistic interaction modeling:
\begin{equation}
\label{eq:loss_fct}
    \mathcal{L} = \mathcal{L}_{\text{plan}} + \mathcal{L}_{\text{wta}} + \alpha \cdot \mathcal{L}_{\text{pred}}.
\end{equation}
Here, $\mathcal{L}_{\text{plan}}$ is the average L2 distance between the planned EV trajectory and ground truth.
$\mathcal{L}_{\text{wta}}$ considers mode selection by assigning a loss of $0$ if the closest mode to the ground truth is selected correctly and else $1$.
$\mathcal{L}_{\text{pred}}$ is the minimal average L2 error for SVs, weighted with $\alpha=0.1$.

\section{RESULTS \& DISCUSSION}

\subsection{Implementation}
All results are gathered with the same architecture using $N=4$ feature update blocks with $Q=4$ relationship categories, a coordinate dimension of $C=D=64$, and $K=6$ trajectory decoders.
Past and future trajectories are encoded as $T_p=4$ and $T_f=6$ positions, which corresponds to $t_p=\qty{1.5}{s}$ and $t_f=\qty{3}{s}$ in the selected dataset, respectively.

\subsection{Dataset}
Since PEP performs joint prediction and planning, we use only multi-vehicle scenes in the official nuScenes prediction split, i.e. 471 training and 136 test scenes~\cite{caesar2020nuscenes}.
These are only 607 of 1000 total scenes.
Route attraction uses the high-level route the driver was supposed to follow during data acquisition, which is provided in the CAN-Bus expansion.

\subsection{Training Setup}
PEP is implemented in PyTorch and has \qty{1.3}{M} trainable parameters when configured as described in A. It is trained over $400$ epochs with batch size 512.
We used the Adam optimizer~\cite{kingma2014adam} with an initial learning rate of \num{5e-4} that decreases with a factor of $0.8$ every other epoch.
On a single GTX 1080Ti training to convergence takes about \qty{1.25}{h}.
\begin{table}[t]
\caption{Planning results on nuScenes}
\label{table:main_results}
\begin{center}
\resizebox{1.0\columnwidth}{!}{
\begin{tabular}{c c c c c c c c c c}
\hline
\multirow{2}{*}{Model} & \multirow{2}{*}{Per} & \multirow{2}{*}{GC} & \multirow{2}{*}{Vel} & \multirow{2}{*}{Acc} & \multirow{2}{*}{Traj} & \multicolumn{2}{c}{L2 (\unit{m})} & \multicolumn{2}{c}{CR (\unit{\%})}\\
 & & & & & & \qty{3}{s} & Avg. & \qty{3}{s} & Avg.\\
\hline
NMP\cite{zeng2019end} & \checkmark & \textbf{-} & \textbf{-} & \textbf{-} & \textbf{-} & 2.31 & \textbf{-} & 1.92 & \textbf{-}\\
SA-NMP\cite{zeng2019end} & \checkmark & \textbf{-} & \textbf{-} & \textbf{-} & \textbf{-} & 2.05 & \textbf{-} & 1.59 & \textbf{-}\\
FF \cite{hu2021safe} & \checkmark & \textbf{-} & \textbf{-} & \textbf{-} & \textbf{-} & 2.54 & 1.43 & 1.07 & 0.43\\
EO \cite{khurana2022differentiable} & \checkmark & \textbf{-} & \textbf{-} & \textbf{-} & \textbf{-} & 2.78 & 1.60 & 0.88 & 0.33\\
\hline
ST-P3 \cite{hu2022st} & \checkmark & \checkmark & \textbf{-} & \textbf{-} & \textbf{-} & 2.90 & 2.11 & 1.27 & 0.71\\
UniAD \cite{hu2023planning} & \checkmark & \checkmark & \textbf{-} & \textbf{-} & \textbf{-} & 1.65 & 1.03 & 0.71 & 0.31\\
DeepEM \cite{chen2023deepemplanner} & \checkmark & \checkmark & \textbf{-} & \textbf{-} & \textbf{-} & 0.73 & 0.48 & 0.36 & 0.19\\
FusionAD \cite{ye2023fusionad} & \checkmark & \checkmark & \checkmark & \checkmark & \checkmark & \textbf{-} & 0.81 & 0.27 & \textbf{0.12}\\
VAD-Tiny \cite{jiang2023vad} & \checkmark & \checkmark & \checkmark & \checkmark & \checkmark & 0.65 & 0.41 & 0.27 & 0.16\\
VAD-Base \cite{jiang2023vad} & \checkmark & \checkmark & \checkmark & \checkmark & \checkmark & 0.60 & 0.37 & \textbf{0.24} & 0.14\\
BEV-Planner++ \cite{li2023ego} & \checkmark & \checkmark & \checkmark & \checkmark & \checkmark & 0.57 & 0.35 & 0.73 & 0.34\\
\hline
AD-MLP-I \cite{zhai2023rethinking} & \textbf{-} & \textbf{-} & \textbf{-} & \textbf{-} & \checkmark & 1.48 & 0.97 & 0.83 & 0.49\\
AD-MLP-II \cite{zhai2023rethinking} & \textbf{-} & \textbf{-} & \checkmark & \checkmark & \checkmark & 0.49 & 0.35 & 0.28 & 0.23\\
AD-MLP-IV \cite{zhai2023rethinking} & \textbf{-} & \checkmark & \checkmark & \checkmark & \checkmark & 0.41 & 0.29 & \textbf{0.24} & 0.19\\
\hline
\textbf{PEP (Ours)} & \textbf{-} & \checkmark & \textbf{-} & \textbf{-} & \checkmark & \textbf{0.32} & \textbf{0.28} & 0.43 & 0.37\\
\hline
\end{tabular}
}
\end{center}
\end{table}
\subsection{Planning}
\begin{figure*}[h]
    \centering
    \includegraphics[width=\linewidth]{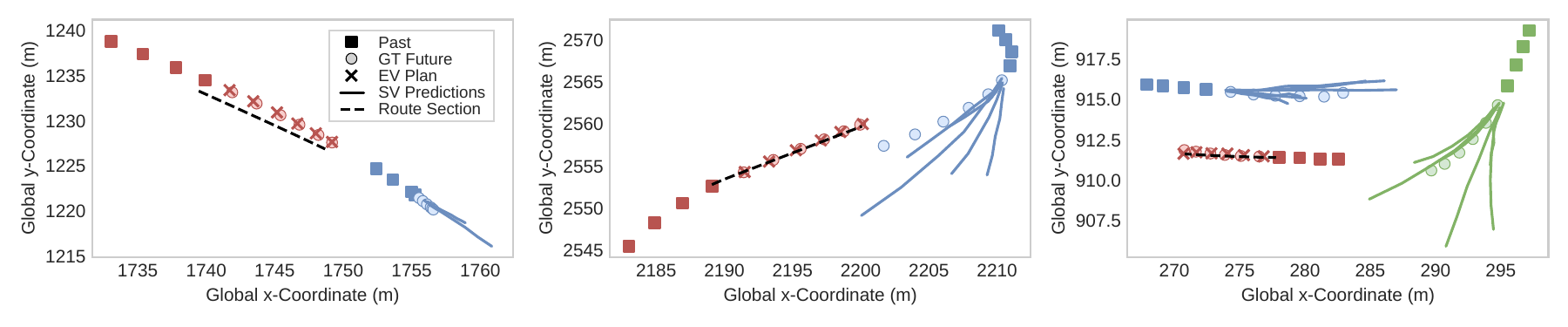}
    \caption{Qualitative results. While the EV (red) uses the route (dashed) for guidance, it does not stick to it (left). Predicting actions of SVs improves EV planning, for example by anticipating SVs to decelerate (blue, left) or to cross the EV lane (blue, middle). Multi-modal predictions help the planner to consider diverse future scenarios (green, right).}
    \label{qualitative_results}
\end{figure*}
Planning performance is evaluated in open loop.
Table~\ref{table:main_results} provides a broad comparison with other methods.
Except for our model, results are taken from~\cite{zhai2023rethinking, chen2023deepemplanner, ye2023fusionad, li2023ego}.
To facilitate an overview, the methods are categorized based on model design criteria.
'Per' indicates that a method uses additional information from perception, 'GC' stands for goal conditioning of the EV, and 'Vel', 'Acc', and 'Traj' encode whether ground truth velocity, acceleration, and trajectory are provided, respectively.
L2 distance between the planned trajectory and ground truth trajectory is used as the main metric.
Additionally, the Collision Rate (CR) is evaluated.\\
PEP achieves the lowest L2 distance at the last planned position, \qty{3}{s} into the future as well as averaged along the trajectory.
Regarding the CR, PEP performs slightly worse than methods, which additionally use ground truth velocity and acceleration as input.
However, the performance is similar to other methods that, like PEP, do not do so.\\
The results suggest that route attraction becomes increasingly beneficial the longer the planning horizon gets.
Compared to SOTA, the L2(\qty{3}{s}) is reduced by \qty{28.1}{\%} while the L2(Avg.) decreases by \qty{3.6}{\%}.
We assume that the L2(Avg.) and CR could be further reduced by incorporating map information under consideration of roto-translation equivariance.
Map information should lead to more accurate interaction modeling, which increases prediction and, thus, planning performance.
Including a map will therefore be the next step to further improve our lightweight map-less approach.
The qualitative results in Fig.~\ref{qualitative_results} showcase how PEP benefits from prediction, route, and multi-modality.

\subsection{Prediction}
Even though prediction is not the primary task of PEP, it leverages joint prediction for realistic interaction modeling when planning the EV trajectory.
During our planning experiments, we measured an SV prediction performance with a minL2(Avg.) of \qty{0.82}{m} and a minL2(\qty{3}{s}) of \qty{0.99}{m}. Considering that no map is available for the SVs, these results are worth mentioning.
In the following, we investigate whether planning really benefits from joint prediction.

\subsection{Ablation}
We present ablation studies for the major design choices of our model.
To assess the impact of SV predictions on EV planning performance, $\mathcal{L}_{\text{pred}}$ is removed from the loss function (c.f. Eq.~\ref{eq:loss_fct}) so that the model is not explicitly trained to predict SVs.
Route ablation is realized by skipping the route attraction module described in Eq.~\ref{eq:route_att_update_rule}.
Finally, we deliberately destroy the SE(2)-equivariance of PEP by not subtracting and re-adding the mean position $\bar{\mathbb{X}}$ during equivariant feature initialization (c.f. Eq.~\ref{eq:equiv_feature_init}).
All networks are trained until convergence.\\
Overall, the ablation experiments show that each component contributes to the planning performance.
Ablating all components at once yields the highest L2 distances and CR(Avg.) but not the highest CR(\qty{3}{s}).
This is explainable by poor behaviors like driving off-road or stopping, which are the consequence of a map-less and route-less approach without explicit prediction.
Such behaviors increase the L2 distance and reduce the CR in an unreasonable way.
Next, we investigate the effect of ablating individual components.
Ablating equivariance results in the highest L2(\qty{3}{s}) and L2(Avg.) increase, which indicates that the model benefits from the prior knowledge on scene symmetry that is integrated by means of SE(2)-equivariance.
Not integrating this knowledge into the model architecture means that the model has to learn it itself, which reduces the sample efficiency and requires model capacity.
Discarding the route also leads to a severe performance decrease as it takes away the only available map information, making the model fully interaction-based.
In contrast, PEP performs only marginally worse when ablating explicit prediction, which is consistent with recent findings that EV information is decisive for open loop planning on nuScenes where interactions play a minor role~\cite{zhai2023rethinking, li2023ego}.
Our results show that prediction is less important than route information and equivariant model design.
Nevertheless, ablating prediction leads to \qty{-10.7}{\%} L2(Avg.) and \qty{-9.4}{\%} L2(\qty{3}{s}) compared to the complete model.
\begin{table}[t]
\caption{Ablations of PEP model}
\label{table:ablations}
\begin{center}
\resizebox{1.0\columnwidth}{!}{
\begin{tabular}{c c c c c c c}
\hline
\multirow{2}{*}{Prediction} & \multirow{2}{*}{Route} & \multirow{2}{*}{Equivariance} & \multicolumn{2}{c}{L2 (\unit{m})} & \multicolumn{2}{c}{CR (\unit{\%})}\\
 & & & \qty{3}{s} & Avg. & \qty{3}{s} & Avg.\\
\hline
\textbf{-} & \textbf{-} & \textbf{-} & 4.94 & 2.88 & 1.33 & 1.79 \\
\checkmark & \checkmark & \textbf{-} & 2.81 & 2.24 & 1.73 & 1.23 \\
\checkmark & \textbf{-} & \checkmark & 1.71 & 1.46 & 1.40 & 0.85 \\
\textbf{-} & \checkmark & \checkmark & 0.35 & 0.31 & 0.48 & 0.42 \\
\checkmark & \checkmark & \checkmark & 0.32 & 0.28 & 0.43 & 0.37 \\
\hline
\end{tabular}
}
\end{center}
\end{table}

\subsection{Equivariance}
\begin{figure}[t]
    \centering
    \includegraphics[width=\linewidth]{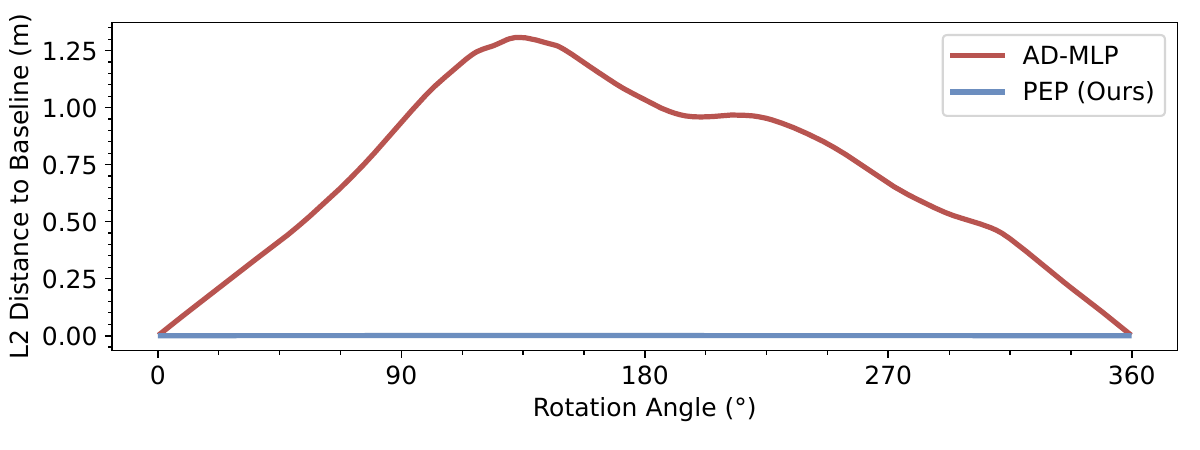}
    \caption{Output stability. Inferred outside the training distribution, our SE(2)-equivariant model guarantees a stable output.}
    \label{output_stability}
\end{figure}

To investigate equivariance, we measure the output stability under input transformations.
To this, the input trajectory and route are rotated by $\theta \in [\qty{1}{\degree}, \qty{2}{\degree}, \dots, \qty{359}{\degree}]$.
Then, the EV trajectory is planned and transformed back into the baseline coordinate system by a rotation of $-\theta$.
The trajectory planned without applying any rotation, i.e. $\theta=0$, serves as the baseline.
For an ideal equivariant model, the L2 distance to the baseline should be zero for all $\theta$.\\
Fig.~\ref{output_stability} depicts the output stability under rotation.
Except for negligible numerical effects from rotation, the L2-distance is constant around zero, demonstrating that PEP is rotation-equivariant.
Repeating the experiment with added random 2D translations confirms the results.
In contrast, AD-MLP~\cite{zhai2023rethinking} which is trained on EV-centered data, is sensitive to input rotations which could, for example, arise from measurement errors.
Especially when designing safety-relevant systems for automated driving, output stability and explainable behavior under input transformations are crucial.

\section{CONCLUSION}
In this work, we have proposed PEP, a simplistic equivariant planning model that integrates prediction and planning in a joint approach.
Our experiments show that PEP achieves state-of-the-art performance in open-loop planning on nuScenes.
Three major design choices contribute to the performance: Joint prediction and planning, our novel route attraction module, and the SE(2)-equivariant network design.
We demonstrate output stability under transformations of the input.
This property of equivariant models can provide safety guarantees and might become an important aspect in the future of automated driving.







\bibliographystyle{IEEEtran}
\bibliography{paper}

\end{document}